\documentclass[11pt]{article}
\usepackage{amsthm,amsmath,amssymb,anysize}
\usepackage{graphicx}

\def\qed{\hbox to 0pt{}\hfill$\rlap{$\sqcap$}\sqcup$}

\usepackage[numbers,sort&compress]{natbib}
 \numberwithin{equation}{subsection}

\begin{document}

\title{\textbf{A New Algorithm based on Extent Bit-array for Computing  Formal Concepts
\footnote{The
authors are supported by   NSF grant of
 Anhui Province(No.1808085MF178), China.} }}
\author{Jianqin Zhou$^{1}$ , Sichun Yang$^{1}$,  Xifeng Wang$^{1}$ and Wanquan Liu$^{2}$ \footnote{Corresponding author.  Email: liuwq63@mail.sysu.edu.cn}\\
\\
\textit{$^{1}$Department of Computer Science,} \\
\textit{Anhui University of   Technology, Ma'anshan 243002,   China}\\
\\
\textit{$^{2}$School of Intelligent Systems Engineering,} \\
\textit{Sun Yat-sen University, Shenzhen 518000,  China}}
\date{ }
\maketitle
\begin{quotation}
\small\noindent

The emergence of Formal Concept Analysis (FCA) as a data analysis technique  has increased the need for developing algorithms which can compute formal concepts quickly. The current efficient algorithms for FCA are variants of the Close-By-One (CbO) algorithm, such as In-Close2, In-Close3 and In-Close4, which are all based on horizontal storage of contexts. In this paper, based on algorithm In-Close4, a new algorithm based on the vertical storage of contexts, called In-Close5, is proposed, which can significantly reduce both the time complexity and space complexity of algorithm In-Close4. Technically, the new algorithm stores both context and extent of a concept as a vertical bit-array, while within In-Close4  algorithm the context is stored only as a horizontal bit-array, which is very slow in finding the intersection of two extent sets.
Experimental results demonstrate that the proposed algorithm is much more effective than In-Close4  algorithm, and it also has a broader scope of applicability
in computing  formal concept in which one can solve the problems that cannot be solved by the In-Close4 algorithm.



\end{quotation}

 \section{Introduction}

Among data analysis techniques, Formal Concept Analysis (FCA) is a useful knowledge representation framework for describing and summarizing data.
As the crucial data structure of FCA, concept lattice is an effective tool for  knowledge discovering, which can depict the generalization and specification between formal concepts in a hierarchical structure. Concept lattice  has been widely used in many areas, such as data mining, machine learning, information retrieval and so on \cite{BDF,Kuz,PIV,PBT, SYW,WW}. The main research contents of concept lattice include lattice construction  \cite{ Outrata,Andrews11, Andrews, Kuznetsov,  Osicka, QWL, QLW, QQW}, rule extraction \cite{LML,LMW,LMWZ,PRM} and lattice reduction  \cite{LMWZ, Kuznetsov2007,Ren}.

A challenging problem in computing these formal concepts is that a typical data set may have a great number of formal concepts.
 It is well known that the number of formal concepts can be increased exponentially in associated with the size of the input context and the problem of determining this number is \#P-complete \cite{ Kuznetsov01}.

In FCbO algorithm, Outrata and Vychodil \cite{ Outrata} introduced an idea in which a concept is closed before its descendants are computed, thus allowing the descendants to fully inherit the attributes of the parent. With the spirit of ‘best-of-breed’ research, this idea was integrated into the  In-Close2 algorithm  \cite{ Andrews11}.

Considering the formal context as a matrix, a row is all the attributes of an object and a column is all the objects of an attribute. Further, all the objects of a formal concept is called extent and all the attributes of a formal concept is called intent.
Within In-Close2, In-Close3 or In-Close4  algorithms \cite{ Andrews11,Andrews, Andrews17}, intents are stored in a linked list tree structure. Extents are stored in a linearised 2-dimensional  array. \textbf{The context is stored as a horizontal bit-array} for optimising for RAM and  cache memory.
This also allows multiple context cells to be processed by a single 32-bit or 64-bit operator.

Suppose that one row of context is
$(1,1,1,1,1,1,1,1$, $0,0,0,0,0,0,0,0,$  $0,0,0,0,$  $0,0,0,0$,  $0,0,0,0$,   $0,0,0,0)$,  it is stored as $31$ in In-Close2, In-Close3 or In-Close4  algorithms, where the first bit means $2^0$, the second bit means $2^1$, the third bit means $2^2$ and so on.
The main shortcoming of these algorithms is that the extent of a concept is not stored as a 32-bit-array (or 64-bit-array), thus they process the intersection of the extent of a concept and a column of context only one object at a time.

A crucial improvement  in our algorithm is that \textbf{both context and extent of a concept are stored as a vertical bit-array} for optimising for RAM and  cache memory, which can significantly reduce both the time complexity and space complexity.
Suppose that one column of context or the extent of a concept is
$(1,1,1,1,1,1,1,1$, $0,0,0,0,0,0,0,0,$  $0,0,0,0,$  $0,0,0,0$,  $0,0,0,0$,   $0,0,0,0)$,  it is stored as $31$ in our algorithm.
Thus multiple context cells are processed by a single 32-bit or 64-bit operator when finding the intersection of the extent of a concept and a column of context.

The second important improvement is the following.
The core procedure in In-Close2 algorithm is ComputeConceptsFrom((A,B),y), which uses a queue of \textbf{local array} \cite{ Andrews11,Andrews}.
In most cases, the local queues are empty, thus the space complexity is not efficient.
In our algorithm, the queue is optimised and used  as \textbf{one global queue}, which would greatly reduce the space
complexity of the core procedure.

This paper illustrates, after a brief description of formal concepts, how formal concepts are computed via In-Close2 algorithm.
 Using a simple example, the  basic recursive process of In-Close2 algorithm is shown, line-by-line.
Using the same notation and style, we present a new variant called the In-Close5. The key differences between the algorithms are then compared to highlight where efficiencies occur.

The paper is organized as follows. In Section \ref{s1}, we review the necessary notions
concerning formal concepts and   their basic properties. In Section \ref{s2}, we study  how formal concepts are computed using In-Close2 algorithm. In Section \ref{s3}, we present the In-Close5 algorithm and give experiment results with In-Close2 algorithm, In-Close4 algorithm and In-Close5 algorithm. Finally   the paper is concluded in Section \ref{s5}.

\section{Basic notions and properties} \label{s1}


In this section, we will review some basic notions and properties of FC involved in this paper. The definitions of a formal context and its operators are given first as follows.

\textbf{Definition 1}. \cite{Ganter} Let $(G, M, I)$ be a formal context, where $G =\{x_1, x_2, \ldots , x_m\}$, $M=\{a_1, a_2, \ldots , a_n\}$, and $I$ is a binary relation between $G$ and $M$. Here each $x_i(i\leq m)$ is called an object, and each $a_j(j \leq n)$ is called an attribute. If an object $g$ has an attribute $m$, we write $gIm$ or $(g, m) \in I$.

\textbf{Definition 2}. \cite{Ganter} Let $(G, M, I)$ be a formal context. For any $A \subseteq G$ and $B \subseteq M$, a pair of positive operators are defined by:

$^* : P(G)\rightarrow P(M)$, $A^* = \{m \in M|\forall g \in A, (g,m) \in I\}$

$^* : P(M)\rightarrow P(G)$, $B^* = \{g \in G | \forall m \in B, (g,m) \in I\}$

Based on the above operators, formal concepts and concept lattices are defined as follows.

\textbf{Definition 3}. \cite{Ganter} Let $(G, M, I)$ be a formal context. For any $A \subseteq G$, $B \subseteq M$, if $A^*=B$ and $B^*=A$, then $(A, B)$ is called a formal concept, where $A$ is called the extent of the formal concept, and $B$ is called the intent of the formal concept.
For any $(X_1, A_1), (X_2, A_2)$, one can define the partial order as follows:

$(X_1, A_1) \leq (X_2, A_2) \Leftrightarrow X_1 \subseteq X_2\Leftrightarrow A_2 \subseteq A_1$

The family of all formal concepts of $(G, M, I)$ is a complete lattice, and it is called a concept lattice and denoted by $L(G, M, I)$.

Let $(G, M, I)$ be a formal context. For any $A_1, A_2, A \subseteq G$, $ B_1, B_2, B \subseteq M$, the following properties hold:

(1) $A_1\subseteq A2 \Rightarrow A^*_2 \subseteq A^*_1$, $\ B_1\subseteq B_2 \Rightarrow B^*_2\subseteq B^*_1$;

(2) $A \subseteq A^{**}$, $\ B \subseteq B^{**}$;

(3) $A^* = A^{***}$, $\ B^* = B^{***}$;

(4) $A \subseteq B^{*}\ \Leftrightarrow\ B \subseteq A^{*}$;

(5) $(A_1\cup A_2)^* = A^*_1\cap A^*_2, \ (B_1\cup B_2)^* = B^*_1\cap B^*_2$;

(6) $(A_1\cap A_2)^* \supseteq   A^*_1\cup A^*_2, \ (B_1\cap B_2)^* \supseteq B^*_1\cup B^*_2$;

Typically a  table of $0$ or $1$ is used to represent a formal context, with $1$s indicating binary relations between objects (rows) and attributes (columns). The following is a simple example of a formal context:

 \begin{table*}[ht]
            \begin{center}
                \begin{normalsize}
                    \caption{Formal context $K=(G, M, I)$}
                    \label{table_tab1}
                    \begin{tabular}
                        {|c|c c  c  c  c|}
                        \hline
                        $G$     & $\ \ \ \ a_1\ \ \ \ \ $       &  $\ \ \ \ a_2\ \ \ \ \ $  &  $\ \ \ \ a_3\ \ \ \ \ $  & $\ \ \ \ a_4\ \ \ \ \ $  &  $\ \ a_5$\ \ \  \\
                       \hline
 1 & 0   & 1  &  1  & 0 &  0  \\

 2 & 1  &  1  & 0 &  0  &  0   \\

 3 & 1   & 0 &  0  &  0  &  0   \\

4 & 0 &  0  &  0  &  0  &   1  \\

5 & 0 &  0  &  0  &  1 &   1  \\

6 & 0 &  0  &  1  &  1 &   1  \\
\hline
                    \end{tabular}
                \end{normalsize}
            \end{center}
        \end{table*}

 The formal concepts in  Table \ref{table_tab1}  can be calculated as  given in the following Table \ref{table_2tab}:

 \begin{table*}[ht]
            \begin{center}
                \begin{scriptsize}
                \caption{Formal concepts in  Table 1}
                    \label{table_2tab}
                    \begin{tabular}
                        {c c c c  }
                        \hline
                             &        &  $C_0=(\{1,2,3,4,5,6\},\emptyset)$  &       \ \ \  \\

                   &    &     &        \\

 $C_4=(\{2,3\},\{a_1\}) $ & $C_3=(\{1,2\},\{a_2\}) $   & $C_2=(\{1,6\},\{a_3\}) $  &  $C_1=(\{4,5,6\},\{a_5\}) $     \\

   &    &     &         \\

  & $C_5=(\{2\},\{a_1, a_2\}) $  &  $C_6=(\{1\},\{a_2, a_3\}) $   & $C_7=(\{5,6\},\{a_4, a_5\}) $      \\

  &    &     &         \\

  &   &    &  $C_8=(\{6\},\{a_3, a_4, a_5\}) $    \\

   &    &     &         \\

 &  &    $C_9=(\emptyset,\{a_1, a_2, a_3 ,a_4, a_5\}) $   &      \\
\hline

                    \end{tabular}
                \end{scriptsize}
            \end{center}
        \end{table*}

  Formal concepts in a  table of $0$ or $1$ can be visualised as closed rectangles of $1$s, where the rows and columns in the rectangle are not necessarily contiguous.
  Suppose we define the cell of the $i$th row and $j$th column as $(i,j)$. Thus in Table \ref{table_tab1}, $(5,4)$, $(5,5)$, $(6,4)$ and $(6,5)$ form the concept $C_7$, and $C_7$ is a rectangle of height $2$ and width $2$. Similarly $(6,3)$, $(6,4)$ and $(6,5)$ form the concept $C_8$, and $C_8$ is a rectangle of height $1$ and width $3$.  $(1,3)$ and $(6,3)$ form the concept $C_2$, and $C_2$ is a rectangle of height $2$ and width $1$, here $(1,3)$ and $(6,3)$ are not contiguous.


In fact, it is not easy to compute the formal concepts given a formal context. Next we will address this problem.

\section{Computation of formal concepts} \label{s2}

A formal concept can be obtained by applying the $^*$ operator to a set of attributes to get its extent, and then applying the $^*$ operator to the extent to get the intent.

For example,  from the context in  Table \ref{table_tab1},  $\{a_4\}^*= \{5,6\}$ and $\{5,6\}^*=\{a_4,a_5\}$. So $(\{5,6\}, \{a_4,a_5\})$ is concept $C_7$ in  Table \ref{table_2tab} .

If this procedure is applied to every possible subset of $\{a_1, a_2, a_3 ,a_4, a_5\}$, then all the concepts in the context can be obtained. However, the number of formal concepts can be exponential in terms of the size of the input context and the problem of determining this number is \#P-complete \cite{ Kuznetsov01}. So an efficient algorithm is crucial and required to compute all the formal concepts in a formal context.

By taking the advantages of algorithm In-Close and algorithm FCbO, In-Close2 is very efficient \cite{Andrews11,Andrews}.
The In-Close2 algorithm, given below, is invoked with an initial $(A,B)=(X,\emptyset)$ and an initial attribute $y=n-1$, where there are $n$ columns in the formal context.

    \begin{figure}[htp] \centering
        \includegraphics[scale=0.6]{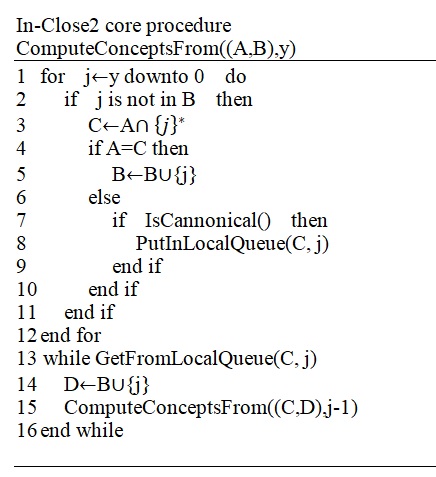}
        \label{figInClose2}
    \end{figure}

\newpage

Line 1 -- Iterate across the context, from starting attribute $y$ down to attribute 0 (the first column).

Line 2 -- Skip attributes already in $B$, as intents now inherit all of their parent’s attributes.

Line 3 -- Form an extent $C$, by intersecting the current extent $A$ with the next column of objects in the context.

Line 4 and Line 5 -- If the extent formed, $C$, equals the extent, $A$, of the concept whose intent is currently being processed, then   add the current attribute $j$ to the intent being processed, $B$.

Line 7 -- Otherwise, check whether $\{j\}^*$ is contained in any new concept in the queue.

Line 8 -- If $\{j\}^*$ is not contained, place the new extent C and the location where it was found, $j$, in a queue for later processing.

Lines 13 -- The queue is processed by obtaining each new extent C and the associated location from the queue.

Line 14 -- Each new partial intent, $D$, inherits all the attributes from its completed parent intent, $B$, along with the attribute, $j$, where its extent was found.

Line 15 -- Call ComputeConceptsFrom to compute child concepts from $j-1$ and to complete the intent $D$.

\

 As the extent of a concept is not stored as a 32-bit-array (or 64-bit-array), thus in Line 3 of ComputeConceptsFrom,  the  algorithm  processes the intersection of the extent of a concept and a column of context only one object at a time, which increases the time complexity of  In-Close2 greatly. This is the main disadvantage of In-Close2 algorithm.

\

For example, apply In-Close2 algorithm to the formal context in Table \ref{table_tab1}, we have results in Table \ref{table_2tab}.
In the first call ComputeConceptsFrom,  $\{a_5\}^*$, $\{a_3\}^*$, $\{a_2\}^*$ and $\{a_1\}^*$   passed through IsCannonical() test.
As $\{a_4\}^*\subset \{a_5\}^*$,  where $\{a_4\}^*=\{5,6\}, \{a_5\}^*=\{4,5,6\}$, so $\{a_4\}^*$   failed IsCannonical() test.

In the second call ComputeConceptsFrom, $\{0\}^*$   passed through IsCannonical() test, we got concept $C_5$ as the child concept of $C_3$.
Similarly, we got concept $C_6$ as the child concept of $C_2$,  $C_7$ as the child concept of $C_1$ and $C_8$ as the child concept of $C_7$.

 By swapping the fourth column and the fifth column in Table \ref{table_tab1}, we have the following Table \ref{table_tab3}.

 \begin{table*}[ht]
            \begin{center}
                \begin{normalsize}
                    \caption{Formal context  }
                    \label{table_tab3}
                    \begin{tabular}
                        {|c|c c  c  c  c|}
                        \hline
                        $G$     & $\ \ \ \ a_1\ \ \ \ \ $       &  $\ \ \ \ a_2\ \ \ \ \ $  &  $\ \ \ \ a_3\ \ \ \ \ $  & $\ \ \ \ a_4\ \ \ \ \ $  &  $\ \ a_5$\ \ \  \\
                       \hline
 1 & 0   & 1  &  1  & 0 &  0  \\

 2 & 1  &  1  & 0 &  0  &  0   \\

 3 & 1   & 0 &  0  &  0  &  0   \\

4 & 0 &  0  &  0  &  1  &  0  \\

5 & 0 &  0  &  0  &  1 &   1  \\

6 & 0 &  0  &  1  &  1 &   1  \\
\hline
                    \end{tabular}
                \end{normalsize}
            \end{center}
        \end{table*}

Apply In-Close2 algorithm to the formal context in Table \ref{table_tab3}, we have results in Table \ref{table_4tab}.

In the first call ComputeConceptsFrom, all $\{a_5\}^*$, $\{a_4\}^*$, $\{a_3\}^*$, $\{a_2\}^*$ and $\{a_1\}^*$   passed through IsCannonical() test.

When call ComputeConceptsFrom with  $A=\{5,6\}$ and $B=\{a_5\}$,   we got $A=A\cap\{a_4\}^*$, where $\{a_4\}^*=\{4,5,6\}$,
 thus $B\leftarrow B\cup \{a_4\}$.

Similarly, we got concept $C_6$ as the child concept of $C_4$,  $C_7$ as the child concept of $C_3$ and $C_8$ as the child concept of $C_1$.

 \begin{table*}[ht]
            \begin{center}
                \begin{tiny}
                \caption{Formal concepts in  Table 3}
                    \label{table_4tab}
                    \begin{tabular}
                        {c c c c c }
                        \hline
                             &        &  $C_0=(\{1,2,3,4,5,6\},\emptyset)$  &     &  \ \ \  \\

                   &    &     &    &    \\
 $C_5= $ & $C_4= $   & $C_3= $  &  $C_2= $    &  $C_1= $  \\
 $(\{2,3\},\{a_1\})$ & $(\{1,2\},\{a_2\})$   & $(\{1,6\},\{a_3\})$  &  $(\{4,5,6\},\{a_4\})$    &  $(\{5,6\},\{a_5,a_4\})$  \\

   &    &     &      &   \\

  & $C_6=(\{2\},\{a_1, a_2\}) $  &  $C_7=(\{1\},\{a_2, a_3\}) $  & & $C_8=(\{6\},\{a_3, a_4, a_5\}) $      \\

   &    &     &     &    \\

 &  &    $C_9=(\emptyset,\{a_1, a_2, a_3 ,a_4, a_5\}) $   &   &   \\
\hline

                    \end{tabular}
                \end{tiny}
            \end{center}
        \end{table*}

In Figure 1, one can see the call tree of ComputeConceptsFrom. From the graph theory, the number of vertices is equal to the number of edges plus one. Here one edge means a ComputeConceptsFrom call  from the queue    and one vertex means an implementation of  ComputeConceptsFrom. One can see that during the first  implementation of  ComputeConceptsFrom,  5 function calls  from the queue are launched. It is obvious that
during the    implementation of  ComputeConceptsFrom,   one function call  from the queue is launched averagely.

Specifically, the local queue of ComputeConceptsFrom is implemented as the following. First int Bchildren[MAX\_COLS] is used to store the location of the attribute  that will spawn new concept. Second   int Cnums[MAX\_COLS] is used
to store the concept number of the spawned concept, where MAX\_COLS$=5,000$. One can see that the efficiency of the local queue is very low.

    \begin{figure}[htp] \centering
        \includegraphics[scale=0.5]{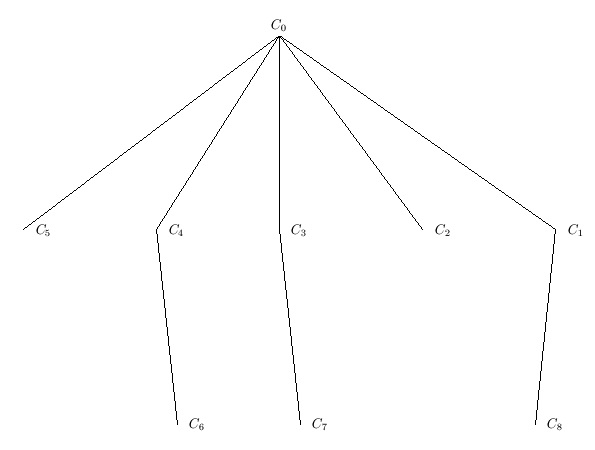}
        \label{figInClose2}
    \end{figure}
    \newpage
\centerline{Figure 1. The call tree of ComputeConceptsFrom}

\

In line 3 of ComputeConceptsFrom,  if the extent formed, $C$, is empty, then store the current attribute $j$, which can be ignored in  concepts of subsequent levels. This is the main improvement from In-Close2 to In-Close3. Further In-Close4 is a 64 bit version, and it can build and output concept trees in JSON format,  where JSON stands for Java Script object notation, is a lightweight data representation method.

\section{In-Close5 algorithm\label{s3}}

Within In-Close2, In-Close3 or In-Close4  algorithms \cite{ Andrews11,Andrews, Andrews17},  extents are stored in a linearised 2-dimensional  array.
A concept of $k$ objects will occupy  $k$ integers. Furthermore, in  ComputeConceptsFrom, the core procedure of  In-Close2,  intersecting the current extent $A$ with the next column of objects in the context is the most time-consuming operation. It inspires us to store both context and extents of
concepts as a vertical bit-array.

 Technically, let $m$ rows in a context be divided into $\lfloor(m-1)/32+1\rfloor$ blocks,  where $\lfloor x\rfloor$ is the largest number that is less than or equal to $x$.  We only store the rows with objects (or nonzero block value) by block number and block value.

For example,  the column $(1,1,1,1,1,1,1,1$, $0,0,0,0$, $0,0,0,0$,  $0,0,0,0$,   $0,0,0,0$,
 $0,0,0,0$,   $0,0,0,0,  0,0,0,0,0,0,0,0)$ is divided into 2 blocks.
 The first block value is $31$, where the first bit means $2^0$, the second bit means $2^1$, the third bit means $2^2$ and so on.
 Thus the column is stored as $\{0, 31\}$, namely block number is $0$ and block value is $31$. We do not store the second block, as the second block value is $0$.

In the case of In-Close4  algorithm, suppose that $(1,1,1,1,1,1,1,1,0,0,0,0,0,0,0,0,$ $0,0,0,0,$ $0,0,0,0,0,0,0,0,0,0,0,0,0,0,0,0,0,0,0,0)$ are the objects of a concept,
then they will stored as $\{0, 1, 2, 3, 4, 5, 6, 7\}$, namely $8$ integers indicate the locations of all objects (or the locations of all $1$s ). 

In the best case of In-Close5, a concept of $k$ objects only occupy  $k/16$ integers, and
one bitwise logic and operation may  process 32 objects when stored as 32-bit integers.
In the worst case of In-Close5, the column $(1,0,0,0, 0,0,0,0,0,0,0,0,0,0,0,0,$  $0,0,0,0,$ $0,0,0,0,  0,0,0,0,   0,0,0,0$ is stored as $\{0, 1\}$,
while In-Close4 the same column is stored as $\{0\}$.

With In-Close2  or In-Close4  algorithm to process mushroom data \cite{Frank}, extents of all concepts will occupy $20,372,788$ bytes of memory.
In contrast, In-Close5  algorithm only needs $4,963,900$ bytes of memory when stored as 32-bit integer,  and $3,505,526$ bytes of memory when stored as 64-bit integer.
From the results above, the space complexity of In-Close5 is much better than that of In-Close2 and In-Close4.

The core procedure in In-Close2 algorithm is ComputeConceptsFrom((A,B),y), which uses a queue of \textbf{local array} \cite{ Andrews11,Andrews}.
Specifically, use int Bchildren[MAX\_COLS] to store the location of the attribute  that will spawn new concept, and int Cnums[MAX\_COLS]
to store the concept number of the spawned concept, where MAX\_COLS$=5,000$.
In fact, the number of times that the function is called  is equal to the number of concepts. Thus in most cases the  queue is empty. This inspires us to link all the local queue together as \textbf{one global queue}, and use the concept number as the index of the queue.
Thus we only need to store the location of the attribute  where the new extent $C$   was found.

The In-Close5 algorithm is presented as the following, which is invoked   with an initial $(A,B)=(X,\emptyset)$, an initial attribute $y=n-1$, where there are $n$ columns in the formal context, and an initial empty Bparent.

    \begin{figure}[htp] \centering
        \includegraphics[scale=0.6]{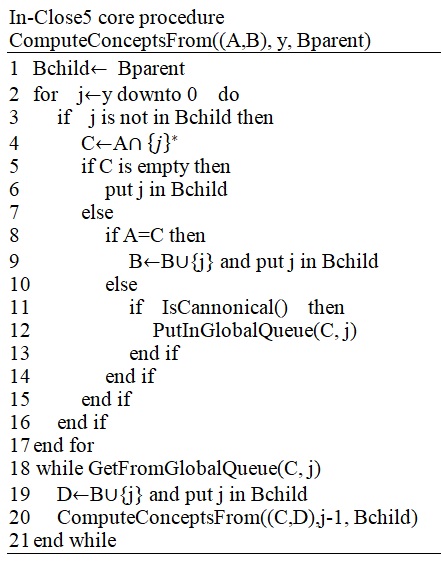}
        \label{figInClose2}
    \end{figure}


Line 1 -- Bparent contains $B$ and the attribute that
  can be ignored in concepts of subsequent levels. The child concept inherits attributes from the parent.

Line 2 -- Iterate across the context, from starting attribute $y$ down to attribute 0 (the first column).

Line 3 -- Skip attributes already in Bchild.

Line 4 -- Form an extent $C$, by intersecting the current extent $A$ with the next column of objects in the context.
It is implemented in C language as the following.

unsigned 	int* Ac = startA[c];						//pointer to start of current extent

unsigned 	int* aptr = startA[highc];					//pointer to start of next extent to be created

			int sizeAc = startA[c+1]-startA[c];			//calculate the size of current extent

			/* iterate across objects in current extent to find them in current  column */

			for(int i = sizeAc/2; i \textgreater  0; i--)$\{$

			\ \ \ \	 if(context0[*Ac][j] \& *(Ac+1))$\{$

			\ \ \ \	 \ \ \ \			*aptr = *Ac;						//add object block number to new extent (intersection)

			\ \ \ \	\ \ \ \			aptr++;

			\ \ \ \	\ \ \ \			*aptr = context0[*Ac][j] \& *(Ac+1);       //add object block value to new extent

			\ \ \ \	\ \ \ \			aptr++;

			\ \ \ \		$\}$

			\ \ \ \		Ac+=2;									//move to next object block

			 $\}$

Line 5 and Line 6 -- If the extent formed, $C$, is empty, then put $j$ in  Bchild, which can be ignored in  concepts of subsequent levels.

Line 8 and Line 9 -- If the extent formed, $C$, equals the extent, $A$, of the concept whose intent is currently being processed, then  add the current attribute $j$ to the intent being processed, $B$ and also put $j$ in  Bchild.

Line 11 -- Otherwise, check whether $\{j\}^*$ is contained in any new concept in the queue.

Line 12 -- If $\{j\}^*$ is not contained, place the   location   $j$  in a global queue for later processing.
It is implemented as the following.

			Bchildren[highc-1] = j;			//note where (attribute column) it was found,
					
					nodeParent[highc] = c;				//note the parent concept number and

					startA[++highc] = aptr;				//note the start of the new extent in A.
			
\

Lines 18 -- The queue is processed by obtaining each new extent $C$ and associated location from the queue.

Line 19 -- Each new partial intent, $D$, inherits all the attributes from its completed parent intent, $B$, along with the attribute, $j$, where its extent was found and attributes that can be ignored in  concepts of subsequent levels.

Line 20 -- Call ComputeConceptsFrom to compute child concepts from $j-1$ and to complete the intent $D$.

Lines 18, Lines 19 and Lines 20 are implemented in C language as the following.

// here numchildrenStart is stored as highc-1 at the beginning of

//ComputeConceptsFrom

for( $i$ = highc-2; $i$ $\geq$ numchildrenStart ; $i$--)$\{$

\ \ \ \	 \ \ \ \			startB[$i$+1] = bptr;						//set the start of the intent in B tree

\ \ \ \	 \ \ \ \    // note that $i+1$ is the number of new extent $C$

\ \ \ \	 \ \ \ \			ComputeConceptsFrom($i+1$, Bchildren[$i$]-1, Bchild);	

$\}$

As both context and extent of a concept are stored as a vertical bit-array,
when form an extent $C$ in Line 4, at most 32 (64)
context cells can be processed by a single 32-bit ( 64-bit)  and operation.
In the case of In-Close3, the extent of a concept is not stored as a 32-bit-array (or 64-bit-array), thus In-Close3 processes context only one cell at a time. 
In Line 12, we only place the location $j$ in a global queue for later
processing,  while In-Close3 has many local empty queues.
The  time complexity and space complexity is greatly reduced, however
the logic structure of In-Close5  algorithm is almost the same as that of In-Close3  algorithm, so please refer to \cite{Andrews} for the correctness of
 In-Close5  algorithm. 

Considering the formal context as a matrix, when transpose the matrix, the concepts of the new matrix should be symmetric to that
of the original matrix. However, there are 8124 columns in transposed mushroom data, thus the depth of recursive calls of ComputeConceptsFrom is greatly increased and so does the complexity.

  For In-Close5  algorithm, with one global queue, it is capable to process transposed mushroom data but with much longer time. In contrast, as local queues use too much memory, In-Close4  algorithm can not process transposed mushroom data.

Some experiments are done  to compare the time complexity of In-Close2  algorithm, In-Close4  algorithm and In-Close5  algorithm.
The   experiment results are given in Table \ref{table_tab5}. Here mushroom data and nursery data are from \cite{Frank}.
The experiments are carried out using a laptop computer with an Intel Core i5-2450M 2.50 GHz processor and 8GB of RAM.

 \begin{table*}[ht]
            \begin{center}
                \begin{normalsize}
                    \caption{Comparison of In-Close2, In-Close4 and In-Close5}
                    \label{table_tab5}
                    \begin{tabular}
                        { c c c  c  c  }
                        \hline
                            & Mushroom       &  Nursery  &  Transposed Mushroom & Transposed Nursery        \ \  \\
                       \hline
$|G|\times |M|$ & $8,124\times 115$   & $12,960\times 30$  &  $ 115 \times  8,124$   & $30\times 12,960$       \\

\#concepts  & 233,101               &  154,055                       & 233,101            &  154,055         \\

 In-Close2 & 0.424                 & 0.123                   &  $\times$               &  $\times$         \\

In-Close4 & 0.388                  &  0.132                    &  $\times$             &  $\times$        \\

In-Close5 & 0.195                  &  0.073                     &  102.536                  &  105.531        \\
\hline
                    \end{tabular}
                \end{normalsize}
            \end{center}
        \end{table*}


 From Table \ref{table_tab5}, one can see that for Mushroom data of $8,124\times 115$, In-Close4 is faster than In-Close2 and In-Close5 is the fastest. For Nursery data of $12,960\times 30$, In-Close4 has no advantage over In-Close2 as 30 is much less than 64.
 As local queues use too much memory, both In-Close2  and In-Close4 can not process transposed mushroom data.

\section{Conclusions and future work\label{s5}}

Within In-Close2, In-Close3 or In-Close4  algorithms, intents are stored in a linked list tree structure and extents are stored in a linearised 2-dimensional   array. The data structure is very simple and effective.
A crucial improvement in our algorithm is that both context and extents of
concepts are stored as a vertical bit-array to optimise for RAM and cache memory, which also significantly reduces   the time for processing extents of
concepts.

Object oriented concept lattice is a more extensive concept lattice \cite{Yao}. It is more difficult to construct object oriented concept lattices.
In the future, we will apply the data structure and technique in these algorithms  to  object oriented concept lattices, 
attribute oriented concept lattices and so on.

%


\vspace{0.2cm}
\begin{thebibliography}{aa}


\bibitem{Andrews11}
S. Andrews,
In-close2, a high performance formal concept miner.
S. Andrews, S. Polovina, R. Hill, B. Akhgar (Eds.), \textit{Conceptual Structures for Discovering Knowledge – Proceedings of the 19th International Conference on Conceptual Structures (ICCS)}, Springer (2011), pp. 50-62

\bibitem{Andrews}
S. Andrews,   A ‘Best-of-Breed’ approach for designing a fast algorithm
for computing fixpoints of Galois Connections, \textit{Information Sciences}, \textbf{295 (20)} (2015) 633--649.

\bibitem{Andrews17}
S. Andrews, In-Close4 Program, 2017,
\textit{https://sourceforge.net/projects/inclose/files/In-Close/}.


\bibitem{BDF} V.G. Blinova, D.A. Dobrynin, V.K. Finn, S.O. Kuznetsov, E.S. Pankratova, Toxicology analysis by means of the jsm-method, \textit{ Bioinformatics.} \textbf{19(10)} (2003) 1201--1207.



\bibitem{Ganter}
B. Ganter, R. Wille, \textit{Formal Concept Analysis:  Mathematical Foundations}, Springer-Verlag, New York, 1999.

\bibitem{Frank}
A. Frank,  A. Asuncion, UCI Machine Learning Repository, 2010, \textit{http://archive.ics.uci.edu/ml}.


\bibitem{Kuznetsov01}
S.O. Kuznetsov,
On computing the size of a lattice and related decision problems,
\textit{Order}, \textbf{18 (4)} (2001) 313-321

\bibitem{Kuznetsov}
S.O. Kuznetsov, S.A. Obiedkov, Comparing performance of algorithms for generating concept lattices,\textit{J. Exp. Theor. Artif. Intell.} \textbf{14(2--3)} (2002) 189--216.

\bibitem{Kuz} S.O. Kuznetsov, Machine learning and formal concept analysis, in: Concept Lattices, \textit{Proceedings of the Second International Conference on Formal Concept Analysis,} ICFCA 2004, Sydney, Australia, February 23-26, 2004,  pp.287-312.


\bibitem{Kuznetsov2007}
S.O. Kuznetsov, S.A. Obiedkov, C. Roth, Reducing the representation complexity of lattice-based taxonomies, in: \textit{Conceptual Structures: Knowledge Architectures for Smart Applications, Proceedings of the 15th International Conference on Conceptual Structures}, ICCS 2007, Sheffield, UK, July 22-27, 2007,  pp.241--254.




\bibitem{LML}
J. Li, C. Mei, Y. Lv, Incomplete decision contexts: approximate concept construction, rule acquisition and knowledge reduction, \textit{Int. J. Approx. Reason}. \textbf{54(1)} (2013) 149--165.


\bibitem{LMW} J. Li, C. Mei, L. Wang, J. Wang, On inference rules in decision formal contexts, \textit{Int. J. Comput. Intell. Syst}. \textbf{8(1)} (2015) 175--186.

\bibitem{LMWZ}J. Li, C. Mei, J. Wang, X. Zhang, Rule-preserved object compression in formal decision contexts using concept lattices, \textit{Knowl.-Based Syst}.\textbf{ 71} (2014) 435--445.

\bibitem{Outrata}
J. Outrata, V. Vychodil,
Fast algorithm for computing fixpoints of Galois connections induced by object-attribute relational data,
\textit{Information Sciences}, \textbf{185 (1)} (2012) 114-127



\bibitem{Osicka}
P. Osicka, Algorithms for computation of concept trilattice of triadic fuzzy context, in:\textit{ Advances in Computational Intelligence -Proceedings of the 14th International Conference on Information Processing and Management of Uncertainty in Knowledge-Based Systems}, IPMU 2012, Catania, Italy, July 9-13, 2012, pp.221-230 (Part III).


\bibitem{PBT}
N. Pasquier, Y. Bastide, R. Taouil, L. Lakhal, Efficient mining of association rules using closed itemset lattices, \textit{Inf. Syst.} \textbf{24(1)} (1999) 25--46.


\bibitem{PIV}
J. Poelmans, D.I. Ignatov, S. Viaene, G. Dedene, S.O. Kuznetsov, Text mining scientific papers: a survey on FCA-based information retrieval research, in:\textit{Advances in Data Mining. Applications and Theoretical Aspects -Proceedings of the 12th Industrial Conference}, ICDM 2012, Berlin, Germany, July 13-20,  2012, pp.273-287.



\bibitem{PRM}Z. Pei, D. Ruan, D. Meng, Z. Liu, Formal concept analysis based on the topology for attributes of a formal context, \textit{Information Sciences},  \textbf{236} (2013) 66--82.




\bibitem{QLW}
J. Qi, W. Liu, L. Wei, Computing the set of concepts through the composition and decomposition of formal contexts, in: \textit{International Conference on Machine Learning and Cybernetics, Proceedings, ICMLC 2012}, Xian, Shaanxi, China, July 15--17,  2012, pp.1326--1332.

\bibitem{QQW}
J. Qi, T. Qian, L. Wei, The connections between three-way and classical concept lattices, \textit{Knowl.-Based Syst}. \textbf{91} (2016) 143--151.


\bibitem{QWL}
J. Qi, L. Wei, Z. Li, A partitional view of concept lattice, in: \textit{Rough Sets, Fuzzy Sets, Data Mining, and Granular Computing, Proceedings of the 10th International Conference}, RSFDGrC 2005, Regina, Canada, August 31--September 3,  2005, pp.74--83 (Part I).



\bibitem{Ren} R. Ren, L. Wei, The attribute reductions of three-way concept lattices, \textit{Knowl.-Based Syst}. \textbf{99} (2016) 92--102.



\bibitem{SYW}
M. Shao, H. Yang, W. Wu, Knowledge reduction in formal fuzzy contexts, \textit{Knowl.-Based Syst.} \textbf{73} (2015) 265--275.

\bibitem{WW}
Q. Wan, L. Wei, Approximate concepts acquisition based on formal contexts, \textit{Knowl.-Based Syst.} \textbf{75} (2015) 78--86.

\bibitem{Yao}
Y. Y. Yao,  Concept lattices in rough set theory,  \textit{Processing Nafips 04 IEEE Meeting of the the Fuzzy Information},
Canada: IEEE,  September 27,2004: 796-801.



\end{thebibliography}
\end{document}